%% file: 0-Main.tex
\documentclass{article}
\usepackage{spconf,amsmath,graphicx,hyperref}

\usepackage{amsfonts}
\usepackage{booktabs}
\usepackage{xcolor}
\usepackage{multirow}
\newcommand{\algname}{CRAFT}

\title{Channel-wise Retrieval for Multivariate Time Series Forecasting}
\name{
Junhyeok Kang\sthanks{Corresponding author.}, 
Jun Seo, Soyeon Park, Sangjun Han, Seohui Bae, Hyeokjun Choe, Soonyoung Lee
}
\address{LG AI Research, Republic of Korea}
\begin{document}
%
\maketitle
\begin{abstract}
Multivariate time series forecasting often struggles to capture long-range dependencies due to fixed lookback windows. Retrieval-augmented forecasting addresses this by retrieving historical segments from memory, but existing approaches rely on a \emph{channel-agnostic} strategy that applies the same references to all variables. This neglects inter-variable heterogeneity, where different channels exhibit distinct periodicities and spectral profiles. We propose \algname{}\,(\underline{C}hannel-wise \underline{r}etrieval-\underline{a}ugmented \underline{f}orecas\underline{t}ing), a novel framework that performs retrieval independently for each channel. To ensure efficiency, \algname{} adopts a two-stage pipeline: a sparse relation graph constructed in the time domain prunes irrelevant candidates, and spectral similarity in the frequency domain ranks references, emphasizing dominant periodic components while suppressing noise. Experiments on seven public benchmarks demonstrate that \algname{} outperforms state-of-the-art forecasting baselines, achieving superior accuracy with practical inference efficiency.
\end{abstract}

%
\begin{keywords}
Multivariate Time Series Forecasting, Retrieval-augmented Generation\,(RAG), Channel-wise Retrieval, Sparse Relation Graph
\end{keywords}
\input{1-Introduction}

\input{2-Related_Work}
\input{3-Method}
\input{4-Evaluation}
\input{5-Conclusion}



\newpage
\bibliographystyle{IEEEbib}
\bibliography{strings,refs}
\appendix
\input{6-Appendix}

\end{document}

%% file: 1-Introduction.tex
\section{Introduction}
\label{sec:introduction}

Multivariate time series forecasting plays a pivotal role in various real-world applications, including demand prediction, traffic management, and weather forecasting.\,\cite{kang2025vardrop, liu2023itransformer, kim2020hi}. A central challenge lies in capturing long-range temporal dependencies, since conventional models rely on fixed-length lookback windows that confine the receptive field to recent history and often miss distant yet informative signals\,\cite{wang2024timemixer, trirat2024universal, zeng2023dlinear}. \emph{Retrieval-augmented forecasting} has recently emerged as a promising remedy\,\cite{han2025raft, ning2025ts, liu2024ratd, lewis2020retrieval, lee2025ratta}. By retrieving relevant historical segments from external memory, models can effectively extend their horizon and incorporate long-range information without enlarging model capacity.

\begin{figure}[t]
    \centering
    \includegraphics[width=\linewidth]{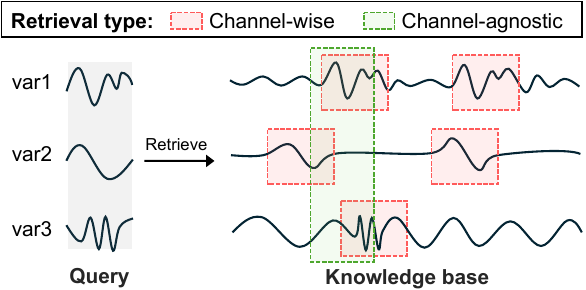}
    \caption{Comparison of retrieval strategies. In channel-agnostic retrieval, all variables use the same retrieved multivariate segment. In channel-wise retrieval\,(ours), each variable independently retrieves its own references, respecting heterogeneity across variables.}
    \label{fig:retrieval_types}
\end{figure}

However, most existing retrieval-augmented methods adopt a \emph{channel-agnostic} strategy in which a single multivariate segment is retrieved from memory and shared across all variables\,\cite{han2025raft, liu2024ratd}. In other words, once a reference segment is selected, every channel is forced to use the same retrieved timeline, regardless of its individual characteristics. This design neglects inter-variable heterogeneity, in which different variables often exhibit distinct characteristics. A reference that is highly informative for one channel may be irrelevant for another, resulting in suboptimal retrieval. Figure~\ref{fig:retrieval_types} illustrates this contrast: \emph{channel-agnostic} retrieval\,(\textcolor{teal}{green}) applies the same multivariate reference across all variables, whereas \emph{channel-wise} retrieval\,(\textcolor{red}{red}) allows each channel to retrieve its own relevant references.

In this regard, we propose \algname{}, a framework that performs \emph{channel-wise} retrieval, where each variable independently retrieves its own historical references. This design explicitly addresses heterogeneity by matching each channel with references that reflect its spectral characteristics, such as dominant periodic patterns. A naive implementation, however, would require every variable to search exhaustively across the entire memory, resulting in prohibitive complexity. To ensure efficiency, we adopt a two-stage retrieval pipeline. In the first stage, a sparse relation graph is constructed in the time domain to precompute inter-variable similarities and restrict each channel to a compact set of related candidates. In the second stage, these candidates are ranked by spectral similarity in the frequency domain, which emphasizes dominant periodic components while suppressing high-frequency noise. This enables variable-specific retrieval at scale, avoiding the expense of exhaustive search.

Our contributions are summarized as follows:
\begin{itemize}
    \item We introduce the channel-wise retrieval paradigm for multivariate time series forecasting, where each variable independently retrieves its own references instead of sharing a common segment. 
    \item We design a two-stage retrieval mechanism for efficient inference: a sparse relation graph prunes unrelated variables, and spectral similarity identifies variable-specific references. 
    \item We validate \algname{} on \emph{seven} public benchmarks, demonstrating superior forecasting performance compared to state-of-the-art baselines. 
\end{itemize}

%% file: 2-Related_Work.tex
\section{Related Work}
\label{sec:related_work}

\textbf{Multivariate Time Series Forecasting.}
Multivariate time series forecasting has been studied extensively, with early approaches based on statistical models such as ARIMA\,\cite{siami2018forecasting} and VAR\,\cite{zivot2006vector}, and more recent advances leveraging deep neural networks including RNNs\,\cite{petnehazi2019recurrent}, CNNs\,\cite{wu2022timesnet}, and Transformer-based architectures\,\cite{wu2021autoformer, nie2023patchtst, zhou2021informer}. While these methods have shown strong performance, they are fundamentally limited by fixed-length lookback windows, which confine the receptive field to recent history and often fail to capture long-range dependencies without significantly increasing model size. 

\noindent\textbf{Retrieval-Augmented Time Series Forecasting.}
Retrieval-augmented forecasting has emerged as a promising alternative, extending context by retrieving relevant historical segments from memory. RAFT\,\cite{han2025raft}, for example, demonstrates that incorporating retrieved trajectories can substantially improve long-horizon forecasting by effectively accessing information beyond the local window. However, existing retrieval-augmented methods\,\cite{iwata2020few, yang2022mqretnn} typically employ a channel-agnostic strategy, where a single retrieved segment is uniformly applied across all variables. This overlooks the heterogeneity among variables—each with its own periodicity, seasonality, and dynamics—and can lead to mismatched retrieval and degraded forecasting performance.

%% file: 3-Method.tex
\section{Method}
\label{sec:method}

\subsection{Problem Definition}
    Given a multivariate time series $\mathbf{X}_{t-T+1:t} \in \mathbb{R}^{T \times C}$, where $T$ is the history length and $C$ is the number of variables, \emph{multivariate time series forecasting} aims to predict the future horizon $\mathbf{Y}_{t+1:t+H} \in \mathbb{R}^{H \times C}$ of length $H$. The prediction is made based on a lookback window $\mathbf{X}_{t-L+1:t} \in \mathbb{R}^{L \times C}$ with $L \ll T$. In the retrieval-augmented paradigm, the model additionally has access to a memory $\mathcal{M} = \{(\mathbf{K}_m, \mathbf{V}_m)\}_{m=1}^M$, where $\mathbf{K}_m \in \mathbb{R}^{L \times C}$ is a past lookback window\,(key) and $\mathbf{V}_m \in \mathbb{R}^{H \times C}$ its future horizon\,(value). At inference, the model retrieves relevant references from $\mathcal{M}$ to extend the available context beyond the lookback window.

\subsection{Overall Framework}

\begin{figure}[t]
    \centering
    \includegraphics[width=\linewidth]{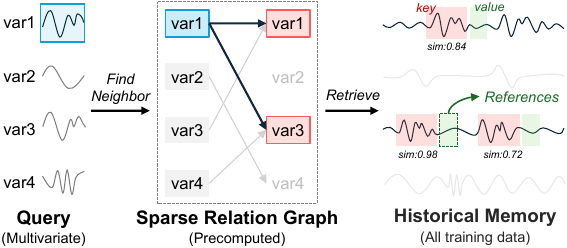}
    \caption{Overall procedure of \algname{}. Each query variable first selects related candidates through a precomputed sparse relation graph, then retrieves variable-specific references from historical memory using spectral similarity.}
    \label{fig:overall_framework}
\end{figure}

    Unlike channel-agnostic approaches that retrieve a single multivariate segment and apply it uniformly to all variables, \algname{} performs \emph{channel-wise retrieval}, where each variable independently selects references tailored to its own characteristics. A direct implementation of this idea would require each variable to compare against the entire memory, which is computationally prohibitive. As illustrated in Fig~\ref{fig:overall_framework}, to make channel-wise retrieval feasible, \algname{} employs a two-stage pipeline: (1) a sparse relation graph in the time domain prunes unrelated variables and restricts retrieval to a compact set of candidates, and (2) spectral similarity in the frequency domain ranks these candidates and identifies variable-specific references. This design enables variable-level retrieval that is both accurate and efficient.

\subsection{Channel-wise Retrieval for Multivariate Time Series}

    \subsubsection{Knowledge Base Construction}
        The first step of \algname{} is to construct a channel-wise knowledge base from the memory $\mathcal{M}$. Naively comparing every variable pair across all samples would be quadratic in $C$ and infeasible. To address this, we precompute inter-variable similarities and retain only the most relevant neighbors for each variable. For each variable $i$, its trajectories across all training samples are concatenated into $\mathbf{z}_i \in \mathbb{R}^{N \cdot L}$, where $N$ is the number of memory samples. Cosine similarity is used to measure inter-variable relations:
        \begin{equation}
        \text{sim}(i,j) = \frac{\mathbf{z}_i^\top \mathbf{z}_j}{\|\mathbf{z}_i\| \, \|\mathbf{z}_j\|}.
        \end{equation}
        Each variable $i$ is connected to its top-$M$ most similar variables, forming a sparse relation graph $\mathcal{G}$. This graph prunes unrelated variables and ensures that retrieval is restricted to semantically close candidates, reducing complexity.
        
        Next, for all candidates in $\mathcal{G}$, we apply fast Fourier transform\,(FFT)\,\cite{brigham1988fast} to obtain spectral representations\,\cite{na2025mitigating}:
        \[
        \mathbf{K}_m^{(c)} = \text{FFT}(\mathbf{K}_m^{(c)})_{:F},
        \]
        where only the first $F$ low-frequency components are retained. This truncation preserves dominant periodic patterns while discarding high-frequency noise and reduces the dimensionality of subsequent similarity computations. The resulting normalized spectra are stored as the channel-wise knowledge base.

    \subsubsection{Retrieval at Inference Time}
        At inference, the model receives a query window $\mathbf{X}_{t-L+1:t}$. For each variable $c$, the sequence $\mathbf{X}_{t-L+1:t}^{(c)}$ is transformed by FFT into its spectrum $\mathbf{q}_c \in \mathbb{C}^{F}$. Only the first $F$ low-frequency components are kept, consistent with the knowledge base. The query $\mathbf{q}_c$ is compared with candidate key spectra $\mathbf{K}_c \in \mathbb{C}^{F \times R}$ retrieved from the knowledge base according to $\mathcal{G}$. Similarity is computed using the normalized complex inner product:
        \begin{equation}
        \text{sim}(\mathbf{q}_c, \mathbf{K}_c) = 
        \frac{\Re\{\mathbf{q}_c \mathbf{K}_c^\ast\}}
        {\|\mathbf{q}_c\| \, \|\mathbf{K}_c\| + \epsilon},
        \end{equation}
        where $\ast$ denotes conjugation and $\Re\{\cdot\}$ the real part. This design emphasizes dominant periodic components while suppressing noise. For each variable $c$, the top-$R$ keys with highest similarity are selected, and their associated values are retrieved as references $\{\mathbf{r}_c^1, \ldots, \mathbf{r}_c^R\}$, with each $\mathbf{r}_c^i \in \mathbb{R}^{H}$. In this way, each variable obtains variable-specific future references rather than being forced to share a single multivariate segment across all channels.
    
\subsection{Forecasting with Retrieved References}
For each variable $c$, the top-$R$ retrieved horizons are first normalized by subtracting the last value\,(offset), projected into the prediction horizon with a linear mapping $f_{\text{retrieved}}(\cdot)$, and then restored by adding back this last value\,\cite{zeng2023dlinear}. The projected results are averaged to form the retrieval-based forecast $\hat{\mathbf{y}}^{\text{retrieval}}_c = \tfrac{1}{R}\sum_{i=1}^R f_{\text{retrieved}}(\mathbf{r}_c^i)$. In parallel, a direct forecast $\hat{\mathbf{Y}}^{\text{direct}}$ is obtained from the input window $\mathbf{X}_{t-L+1:t}$ using a forecasting model $f_{\text{direct}}$. The final prediction is then computed as $\hat{\mathbf{Y}} = \hat{\mathbf{Y}}^{\text{direct}} + \alpha \cdot \hat{\mathbf{Y}}^{\text{retrieval}}$, where $\alpha$ is a fixed coefficient. Because this refinement is formulated as a simple additive correction, the retrieval component can be seamlessly integrated into any forecaster $\hat{\mathbf{Y}}^{\text{direct}}$ without altering its architecture. In this study, we adopt a two-layer MLP\,(multi-layer perceptron) integrated with the normalization technique of NLinear\,\cite{zeng2023dlinear} to alleviate the distribution shifts in time series data. The normalization strategy consists of subtracting the last value of the input window $\mathbf{X}_{t}$ from the input window and adding it to the prediction of forecaster $f_{\text{direct}}$.

\subsection{Complexity Analysis of \algname{}}
    Naive channel-wise retrieval requires $\mathcal{O}(C^2 \cdot |\mathcal{M}|)$ similarity computations, which is infeasible for large $C$. Our design reduces this cost in two ways. First, the sparse relation graph restricts each variable to $M \ll C$ candidates, avoiding quadratic scaling. Second, the frequency cutoff reduces the dimensionality of similarity computation from $L$ to $F \ll L$. The overall complexity becomes $\mathcal{O}(C \cdot M \cdot F)$, which scales linearly with the number of variables $C$. Since both $M$ and $F$ are small relative to $C$ and $L$, this design ensures that channel-wise retrieval remains efficient.

%% file: 4-Evaluation.tex
\section{Evaluation}
\label{sec:evaluation}

\subsection{Experimental Setup}
\label{subsec:experimental_setup}

    \textbf{Datasets.} 
    We evaluate \algname{} on seven widely used multivariate time series benchmarks: ETTh1, ETTh2, ETTm1, ETTm2, Electricity\,(ECL), Traffic, and Weather\,\cite{wu2021autoformer}. These datasets span diverse domains such as energy consumption, transportation, and meteorology, and have been widely adopted as standard benchmarks in the literature\,\cite{han2025raft, wu2022timesnet}. For each dataset, we follow the common evaluation protocol, setting forecasting horizons $H \in \{96,192,336,720\}$ and reporting the average performance.
        
    \noindent \textbf{Baselines.}
    We compare \algname{} against \emph{eight} state-of-the-art multivariate time series forecasting methods: RAFT\,\cite{han2025raft}, TimeMixer\,\cite{wang2024timemixer}, PatchTST\,\cite{nie2023patchtst}, TimesNet\,\cite{wu2022timesnet}, MICN\,\cite{wang2023micn}, DLinear\,\cite{zeng2023dlinear}, Stationary\,\cite{liu2022nonstationary}, and Autoformer\,\cite{wu2021autoformer}.
    
    \noindent \textbf{Evaluation Metrics.} 
    We report mean squared error\,(MSE) and mean absolute error\,(MAE), following prior work\,\cite{han2025raft}.
    
    \noindent \textbf{Implementation Details.} 
    We set the lookback length to $L=720$ across all datasets. In the sparse relation graph, the number of candidates $M$ is set to 3. During retrieval, the number of references $k$ is set to 1, selected by spectral similarity over the first $F=36$ frequency components\,(5\% of the spectrum). The weighting coefficient $\alpha$ is set to 0.001. \algname{} is trained with Adam using a learning rate of 0.001 and a batch size of 32 in PyTorch 2.0.1, and evaluated on an RTX 3090 GPU.

\subsection{Forecasting Performance Comparison}
\input{Tables/forecasting_performance}
    Table~\ref{tab:forecasting_performance} reports the performance comparison of \algname{} against state-of-the-art baselines on seven public benchmark datasets. 
    We follow the common evaluation protocol by setting the forecasting horizon $H \in \{96,192,336,720\}$ and reporting the average results in terms of MSE and MAE\,\cite{han2025raft}. 
    As shown in the Table ~\ref{tab:forecasting_performance}, \algname{} achieves the best average performance across all benchmarks, reducing both MSE and MAE compared to strong recent baselines. 
    These results demonstrate the effectiveness of channel-wise retrieval: by allowing each variable to reference its own relevant history, \algname{} avoids the mismatches inherent in channel-agnostic designs and achieves comparable forecasting accuracy.

\subsection{Efficiency of \algname{}}
\begin{figure}[t]
        \centering
        \includegraphics[width=\linewidth]{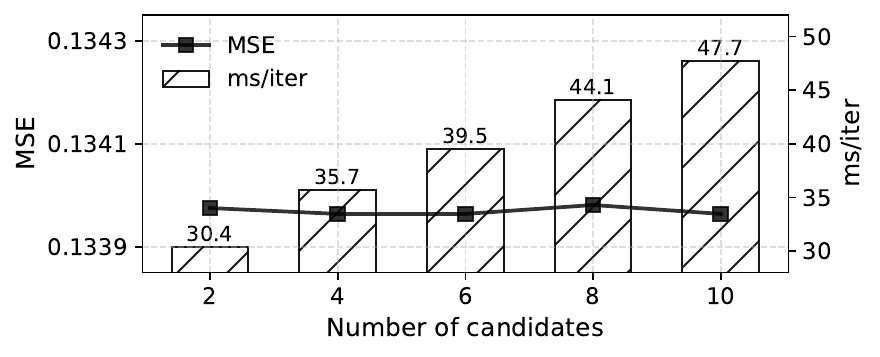}
        \caption{Impact of varying the number of candidate neighbors $M$ in the sparse relation graph on the Electricity dataset with forecast horizon $H=96$. Reducing $M$ lowers inference cost while maintaining multivariate time series forecasting accuracy, showing the feasibility of real-time applications.}
        \label{fig:varying_n_candidates}
    \end{figure}

    We evaluate the efficiency of the sparse relation graph by varying the number of candidate neighbors $M$ for each variable. Figure~\ref{fig:varying_n_candidates} reports results on the Electricity dataset with lookback $L=720$ and horizon $H=96$. As $M$ increases, inference time grows steadily, while forecasting accuracy improves only marginally. Notably, \algname{} maintains strong accuracy with as few as $M=3$ neighbors among all 321 variables, confirming that restricting retrieval to a small candidate set is sufficient for effective forecasting. Moreover, the overall inference time per batch iteration\,(batch size is set to 32) is only about 0.03 seconds, corresponding to more than 1,000 time series instances per second. This demonstrates that the proposed design is not only accurate but also efficient enough for real-time forecasting services.

\subsection{Qualitative Analysis through Visualization}
\begin{figure}[t]
    \centering
    \includegraphics[width=\linewidth]{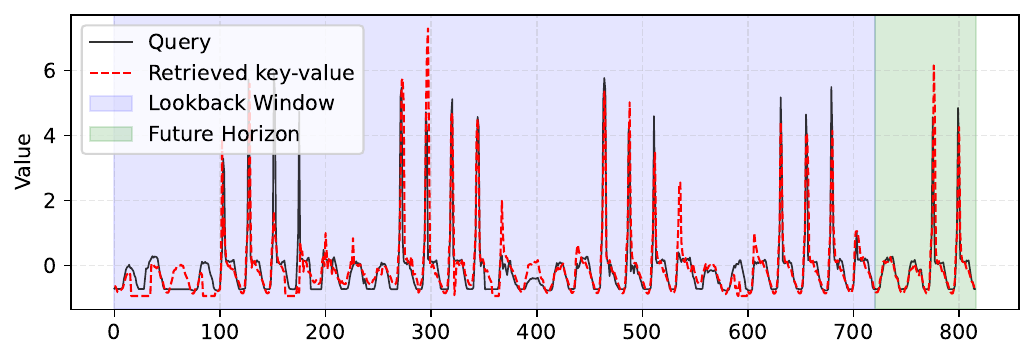}
    \caption{Visualization on the Traffic dataset with $L=720$ and $H=96$. The retrieved key-value pair\,(\textbf{\textcolor{red}{red}}) closely matches the ground-truth time series\,(\textbf{black}), showing that retrieval provides useful references that directly benefit forecasting.}
    \label{fig:retrieved_example}
\end{figure}

To demonstrate the effectiveness of \algname{}, we visualize an example time-series and its retrieved time-series from the Traffic dataset in Figure~\ref{fig:retrieved_example}. The case considers one variable with lookback length $L=720$ and forecasting horizon $H=96$. The figure compares the true future horizon with the retrieved key-value pair obtained from memory. Strikingly, the retrieved sequence matches the ground truth over a large portion of the horizon, capturing both the dominant periodic trend and finer local fluctuations. This alignment illustrates how retrieval provides informative signals that complement the limited lookback window, thereby enhancing the forecast performance. The result highlights that when relevant historical references are retrieved, they can substantially improve forecasting accuracy in practice. 

%% file: Tables/forecasting_performance.tex
\begin{table*}[t]
    \setlength{\tabcolsep}{3.4pt}
    \centering
    \resizebox{1\linewidth}{!}{
        \begin{tabular}{c|cc|cc|cc|cc|cc|cc|cc|cc|cc}
        \toprule
        Model &  \multicolumn{2}{c|}{\textbf{\algname{}}} & \multicolumn{2}{c|}{RAFT} & \multicolumn{2}{c|}{TimeMixer} & \multicolumn{2}{c|}{PatchTST} & \multicolumn{2}{c|}{TimesNet} & \multicolumn{2}{c|}{MICN}& \multicolumn{2}{c|}{DLinear} & \multicolumn{2}{c|}{Stationary}& \multicolumn{2}{c}{Autoformer}\\ 
         &  \multicolumn{2}{c|}{(Ours)} & \multicolumn{2}{c|}{\small(ICML 2025)} & \multicolumn{2}{c|}{\small(ICLR 2024)} & \multicolumn{2}{c|}{\small(ICLR 2023)} & \multicolumn{2}{c|}{\small(ICLR 2023)} & \multicolumn{2}{c|}{\small(ICLR 2023)}& \multicolumn{2}{c|}{\small(AAAI 2023)} & \multicolumn{2}{c|}{\small(NeurIPS 2022)}& \multicolumn{2}{c}{\small(NeurIPS 2021)}\\ 
        
        \midrule
        
        Metric & MSE & MAE & MSE & MAE & MSE & MAE & MSE & MAE & MSE & MAE  & MSE &MAE  & MSE & MAE  & MSE &MAE  & MSE & MAE \\
        \midrule
        ETTh1 &  0.420&0.434& 0.420& 0.436&0.447& 0.440& 0.516&0.484& 0.495& 0.450 & 0.475&0.481&0.461&0.458  & 0.570&0.537& 0.496 & 0.487\\
        ETTh2 &  0.340&0.390& 0.359& 0.409&0.364& 0.395& 0.391&0.412& 0.414& 0.427 & 0.574&0.532&0.563&0.519  & 0.526&0.516& 0.450 & 0.459\\
        
        ETTm1&  0.360&0.383& 0.348&0.378&0.381& 0.396& 0.406&0.409& 0.400& 0.406 & 0.423&0.422&0.404&0.408  & 0.481&0.456& 0.588 & 0.517\\
        ETTm2&  0.250&0.314& 0.254&0.320&0.275& 0.323& 0.290&0.334& 0.291& 0.333 & 0.353&0.402&0.354&0.402  & 0.306&0.347& 0.327 & 0.371\\
        ECL&  0.163&0.257& 0.160&0.259&0.182& 0.273& 0.216&0.318& 0.193& 0.304 & 0.196&0.309&0.225&0.319  & 0.193&0.296& 0.227 & 0.364\\
        Traffic& 0.412&0.284& 0.401&0.282&0.484& 0.298& 0.529&0.341& 0.620& 0.336 & 0.593&0.356&0.625&0.383  & 0.624&0.340& 0.628&0.379\\
        Weather&  0.240&0.281& 0.241&0.286&0.240& 0.272& 0.265&0.286& 0.251& 0.294 & 0.268&0.321&0.265&0.315  & 0.288&0.314& 0.338&0.382\\
    
        \midrule
        \emph{Average} & \textbf{0.312}& \textbf{0.334}& 0.312& 0.339& 0.339& 0.342& 0.373& 0.369& 0.381& 0.364  & 0.412&0.403& 0.414 & 0.401  & 0.427&0.401& 0.436 & 0.423\\
        \bottomrule
        
        \end{tabular}
    }
    \caption{Performance comparison of multivariate time series forecasting with state-of-the-art baselines on the public benchmark datasets. We set forecasting horizon $H$ $\in \{96, 192, 336, 720\}$ for all methods and report the average. The results are mostly taken from previous studies\,\cite{han2025raft}. Please refer to Table~\ref{tab:forecasting_performance_full_extended} in Appendix~\ref{sec:appendix-full_exp_result} for the full results across all forecasting horizons.}
    \label{tab:forecasting_performance}
\end{table*}

%% file: 5-Conclusion.tex
\section{Conclusion}
\label{sec:conclusion}

In this paper, we introduced \algname{}, a channel-wise retrieval-augmented forecasting framework that moves beyond the limitations of channel-agnostic designs. By allowing each variable to access its own historical references and constraining retrieval through a sparse relation graph and spectral similarity, \algname{} provides variable-specific context without incurring prohibitive cost. Experiments on seven public benchmarks show consistent improvements over state-of-the-art methods, underscoring the practical value of retrieval-based approaches for multivariate time series forecasting.

%% file: 6-Appendix.tex
\clearpage 
\onecolumn
\appendix

\section{Full Experimental Results}
\label{sec:appendix-full_exp_result}

\input{Tables/forecasting_performance_full}

%% file: Tables/forecasting_performance_full.tex
\begin{table*}[!ht]
\caption{Full performance comparison of multivariate time series forecasting with state-of-the-art baselines on the public benchmark datasets. We set forecasting horizon $H$ $\in \{96, 192, 336, 720\}$ for all methods. The results are mostly taken from previous studies\,\cite{han2025raft}.}
\setlength{\tabcolsep}{3.4pt}
\small
\centering
\resizebox{1\linewidth}{!}{
    \begin{tabular}{c|c|cc|cc|cc|cc|cc|cc|cc|cc|cc}
        \toprule
        \multicolumn{2}{c|}{\multirow{2}{*}{Models}} & \multicolumn{2}{c|}{\textbf{\algname{}}} & \multicolumn{2}{c|}{RAFT} & \multicolumn{2}{c|}{TimeMixer} & \multicolumn{2}{c|}{PatchTST} & \multicolumn{2}{c|}{TimesNet} & \multicolumn{2}{c|}{MICN}& \multicolumn{2}{c|}{DLinear} & \multicolumn{2}{c|}{Stationary}& \multicolumn{2}{c}{Autoformer}\\ 
        \multicolumn{2}{c|}{} & \multicolumn{2}{c|}{(Ours)} & \multicolumn{2}{c|}{\small(ICML'25)} & \multicolumn{2}{c|}{\small(ICLR'24)} & \multicolumn{2}{c|}{\small(ICLR'23)} & \multicolumn{2}{c|}{\small(ICLR'23)} & \multicolumn{2}{c|}{\small(ICLR'23)}& \multicolumn{2}{c|}{\small(AAAI'23)} & \multicolumn{2}{c|}{\small(NeurIPS'22)}& \multicolumn{2}{c}{\small(NeurIPS'21)}\\
        \midrule
        \multicolumn{2}{c|}{Metric} & MSE & MAE & MSE & MAE & MSE & MAE & MSE & MAE & MSE & MAE  & MSE & MAE  & MSE & MAE  & MSE & MAE  & MSE & MAE \\
        \midrule
        
        \multirow{5}{*}{ETTh1} 
        & 96  & 0.371 & 0.398 & 0.367 & 0.397 & 0.375 & 0.400 & 0.460 & 0.447 & 0.384 & 0.402 & 0.426 & 0.446 & 0.397 & 0.412 & 0.513 & 0.491 & 0.449 & 0.459 \\
        & 192 & 0.411 & 0.422 & 0.411 & 0.427 & 0.429 & 0.421 & 0.512 & 0.477 & 0.436 & 0.429 & 0.454 & 0.464 & 0.446 & 0.441 & 0.534 & 0.504 & 0.500 & 0.482 \\
        & 336 & 0.443 & 0.443 & 0.436 & 0.442 & 0.484 & 0.458 & 0.546 & 0.496 & 0.638 & 0.469 & 0.493 & 0.487 & 0.489 & 0.467 & 0.588 & 0.535 & 0.521 & 0.496 \\
        & 720 & 0.456 & 0.471 & 0.467 & 0.478 & 0.498 & 0.482 & 0.544 & 0.517 & 0.521 & 0.500 & 0.526 & 0.526 & 0.513 & 0.510 & 0.643 & 0.616 & 0.514 & 0.512 \\
        \cmidrule{2-20}
        & Avg & 0.420 & 0.434 & 0.420 & 0.436 & 0.447 & 0.440 & 0.516 & 0.484 & 0.495 & 0.450 & 0.475 & 0.481 & 0.461 & 0.458 & 0.570 & 0.537 & 0.496 & 0.487 \\
        \midrule
        
        \multirow{5}{*}{ETTh2} 
        & 96  & 0.272 & 0.336 & 0.276 & 0.344 & 0.289 & 0.341 & 0.308 & 0.355 & 0.340 & 0.374 & 0.372 & 0.424 & 0.340 & 0.394 & 0.476 & 0.458 & 0.346 & 0.388 \\
        & 192 & 0.333 & 0.377 & 0.347 & 0.393 & 0.372 & 0.392 & 0.393 & 0.405 & 0.402 & 0.414 & 0.492 & 0.492 & 0.482 & 0.479 & 0.512 & 0.493 & 0.456 & 0.452 \\
        & 336 & 0.358 & 0.403 & 0.376 & 0.425 & 0.386 & 0.414 & 0.427 & 0.436 & 0.452 & 0.452 & 0.607 & 0.555 & 0.591 & 0.541 & 0.552 & 0.551 & 0.482 & 0.486 \\
        & 720 & 0.396 & 0.442 & 0.436 & 0.473 & 0.412 & 0.434 & 0.436 & 0.450 & 0.462 & 0.468 & 0.824 & 0.655 & 0.839 & 0.661 & 0.562 & 0.560 & 0.515 & 0.511 \\
        \cmidrule{2-20}
        & Avg & 0.340 & 0.390 & 0.359 & 0.409 & 0.364 & 0.395 & 0.391 & 0.412 & 0.414 & 0.427 & 0.574 & 0.532 & 0.563 & 0.519 & 0.526 & 0.516 & 0.450 & 0.459 \\
        \midrule
        
        \multirow{5}{*}{ETTm1} 
        & 96  & 0.310 & 0.354 & 0.302 & 0.349 & 0.320 & 0.357 & 0.352 & 0.374 & 0.338 & 0.375 & 0.365 & 0.387 & 0.346 & 0.374 & 0.386 & 0.398 & 0.505 & 0.475 \\
        & 192 & 0.337 & 0.368 & 0.329 & 0.367 & 0.361 & 0.381 & 0.390 & 0.393 & 0.374 & 0.387 & 0.403 & 0.408 & 0.382 & 0.391 & 0.459 & 0.444 & 0.553 & 0.496 \\
        & 336 & 0.367 & 0.386 & 0.355 & 0.383 & 0.390 & 0.404 & 0.421 & 0.414 & 0.410 & 0.411 & 0.436 & 0.431 & 0.415 & 0.415 & 0.495 & 0.464 & 0.621 & 0.537 \\
        & 720 & 0.426 & 0.423 & 0.406 & 0.413 & 0.454 & 0.441 & 0.462 & 0.449 & 0.478 & 0.450 & 0.489 & 0.462 & 0.473 & 0.451 & 0.585 & 0.516 & 0.671 & 0.561 \\
        \cmidrule{2-20}
        & Avg & 0.360 & 0.383 & 0.348 & 0.378 & 0.381 & 0.396 & 0.406 & 0.409 & 0.400 & 0.406 & 0.423 & 0.422 & 0.404 & 0.408 & 0.481 & 0.456 & 0.588 & 0.517 \\
        \midrule
        
        \multirow{5}{*}{ETTm2} 
        & 96  & 0.162 & 0.253 & 0.164 & 0.256 & 0.175 & 0.258 & 0.183 & 0.270 & 0.187 & 0.267 & 0.197 & 0.296 & 0.193 & 0.293 & 0.192 & 0.274 & 0.255 & 0.339 \\
        & 192 & 0.217 & 0.293 & 0.219 & 0.296 & 0.237 & 0.299 & 0.255 & 0.314 & 0.249 & 0.309 & 0.284 & 0.361 & 0.284 & 0.361 & 0.280 & 0.339 & 0.281 & 0.340 \\
        & 336 & 0.270 & 0.328 & 0.275 & 0.336 & 0.298 & 0.340 & 0.309 & 0.347 & 0.321 & 0.351 & 0.381 & 0.429 & 0.382 & 0.429 & 0.334 & 0.361 & 0.339 & 0.372 \\
        & 720 & 0.351 & 0.381 & 0.359 & 0.392 & 0.391 & 0.396 & 0.412 & 0.404 & 0.408 & 0.403 & 0.549 & 0.522 & 0.558 & 0.525 & 0.417 & 0.413 & 0.433 & 0.432 \\
        \cmidrule{2-20}
        & Avg & 0.250 & 0.314 & 0.254 & 0.320 & 0.275 & 0.323 & 0.290 & 0.334 & 0.291 & 0.333 & 0.353 & 0.402 & 0.354 & 0.402 & 0.306 & 0.347 & 0.327 & 0.371 \\
        \midrule
        
        \multirow{5}{*}{ECL}
        & 96  & 0.134 & 0.230 & 0.133 & 0.232 & 0.153 & 0.247 & 0.190 & 0.296 & 0.168 & 0.272 & 0.180 & 0.293 & 0.210 & 0.302 & 0.169 & 0.273 & 0.201 & 0.317 \\
        & 192 & 0.150 & 0.245 & 0.149 & 0.247 & 0.166 & 0.256 & 0.199 & 0.304 & 0.184 & 0.289 & 0.189 & 0.302 & 0.210 & 0.305 & 0.182 & 0.286 & 0.222 & 0.334 \\
        & 336 & 0.165 & 0.260 & 0.161 & 0.259 & 0.185 & 0.277 & 0.217 & 0.319 & 0.198 & 0.300 & 0.198 & 0.312 & 0.223 & 0.319 & 0.200 & 0.304 & 0.231 & 0.338 \\
        & 720 & 0.204 & 0.293 & 0.197 & 0.297 & 0.225 & 0.310 & 0.258 & 0.352 & 0.220 & 0.320 & 0.217 & 0.330 & 0.258 & 0.350 & 0.222 & 0.321 & 0.254 & 0.361 \\
        \cmidrule{2-20}
        & Avg & 0.163 & 0.257 & 0.160 & 0.259 & 0.182 & 0.273 & 0.216 & 0.318 & 0.193 & 0.304 & 0.196 & 0.309 & 0.225 & 0.319 & 0.193 & 0.296 & 0.227 & 0.364 \\
        \midrule
        
        \multirow{5}{*}{Traffic}
        & 96  & 0.387 & 0.273 & 0.378 & 0.273 & 0.462 & 0.285 & 0.526 & 0.347 & 0.593 & 0.321 & 0.577 & 0.350 & 0.650 & 0.396 & 0.612 & 0.338 & 0.613 & 0.388 \\
        & 192 & 0.401 & 0.282 & 0.391 & 0.277 & 0.473 & 0.296 & 0.522 & 0.332 & 0.617 & 0.336 & 0.589 & 0.356 & 0.598 & 0.370 & 0.613 & 0.340 & 0.616 & 0.382 \\
        & 336 & 0.411 & 0.281 & 0.402 & 0.282 & 0.498 & 0.296 & 0.517 & 0.334 & 0.629 & 0.336 & 0.594 & 0.358 & 0.605 & 0.373 & 0.618 & 0.328 & 0.622 & 0.337 \\
        & 720 & 0.448 & 0.301 & 0.434 & 0.297 & 0.506 & 0.313 & 0.552 & 0.352 & 0.640 & 0.350 & 0.613 & 0.361 & 0.645 & 0.394 & 0.653 & 0.355 & 0.660 & 0.408 \\
        \cmidrule{2-20}
        & Avg & 0.412 & 0.284 & 0.401 & 0.282 & 0.484 & 0.298 & 0.529 & 0.341 & 0.620 & 0.336 & 0.593 & 0.356 & 0.625 & 0.383 & 0.624 & 0.340 & 0.628 & 0.379 \\
        \midrule
        
        \multirow{5}{*}{Weather}
        & 96  & 0.167 & 0.221 & 0.165 & 0.222 & 0.163 & 0.209 & 0.186 & 0.227 & 0.172 & 0.220 & 0.198 & 0.261 & 0.195 & 0.252 & 0.173 & 0.223 & 0.266 & 0.336 \\
        & 192 & 0.212 & 0.262 & 0.211 & 0.264 & 0.208 & 0.250 & 0.234 & 0.265 & 0.219 & 0.261 & 0.239 & 0.299 & 0.237 & 0.295 & 0.245 & 0.285 & 0.307 & 0.367 \\
        & 336 & 0.258 & 0.297 & 0.260 & 0.302 & 0.251 & 0.287 & 0.284 & 0.301 & 0.280 & 0.306 & 0.285 & 0.336 & 0.282 & 0.331 & 0.321 & 0.338 & 0.359 & 0.395 \\
        & 720 & 0.322 & 0.343 & 0.327 & 0.355 & 0.339 & 0.341 & 0.356 & 0.349 & 0.365 & 0.359 & 0.351 & 0.388 & 0.345 & 0.382 & 0.414 & 0.410 & 0.419 & 0.428 \\
        \cmidrule{2-20}
        & Avg & 0.240 & 0.281 & 0.241 & 0.286 & 0.240 & 0.272 & 0.265 & 0.286 & 0.251 & 0.294 & 0.268 & 0.321 & 0.265 & 0.315 & 0.288 & 0.314 & 0.338 & 0.382 \\
        \midrule
        
        \multicolumn{2}{c|}{\emph{Overall}} & \textbf{0.312} & \textbf{0.334} & 0.312 & 0.339 & 0.339 & 0.342 & 0.373 & 0.369 & 0.381 & 0.364 & 0.412 & 0.403 & 0.414 & 0.401 & 0.427 & 0.401 & 0.436 & 0.423 \\
        \bottomrule
        
    \end{tabular}
}
\label{tab:forecasting_performance_full_extended}
\end{table*}